\documentclass[letterpaper]{article} 
\usepackage{aaai25}  
\usepackage{times}  
\usepackage{helvet}  
\usepackage{courier}  
\usepackage[hyphens]{url}  
\usepackage{graphicx} 
\usepackage{amsmath}
\usepackage{natbib}  
\usepackage{caption} 

\usepackage{algorithm}
\usepackage{algorithmic}

\usepackage{times}  
\usepackage{helvet}  
\usepackage{courier}  
\usepackage[hyphens]{url}  
\usepackage{graphicx} 
\usepackage{colortbl}
\usepackage[table]{xcolor}
\usepackage{natbib}  
\usepackage{arydshln}
\usepackage{booktabs}
\usepackage{caption} 
\usepackage{cleveref}   
\usepackage{algorithm}
\usepackage{algorithmic}
\usepackage{newfloat}
\usepackage{listings}
\DeclareCaptionStyle{ruled}{labelfont=normalfont,labelsep=colon,strut=off} %
\usepackage{newfloat}
\usepackage{listings}

\DeclareCaptionStyle{ruled}{labelfont=normalfont,labelsep=colon,strut=off} 
\floatstyle{ruled}
\newfloat{listing}{tb}{lst}{}
\floatname{listing}{Listing}
\title{TNCSE: Tensor's Norm Constraints for Unsupervised Contrastive Learning of Sentence Embeddings}
\author{
    Tianyu Zong\textsuperscript{\rm 1}, Bingkang Shi\textsuperscript{\rm 2}, Hongzhu Yi\textsuperscript{\rm 1}, Jungang Xu\textsuperscript{\rm 1}\thanks{Corresponding author.}
}
\affiliations{
    \textsuperscript{\rm 1}School of Computer Science and Technology, University of Chinese Academy of Sciences\\

    \textsuperscript{\rm 2}School of Cyber Security, University of Chinese Academy of Sciences\\

    \{zongtianyu20, shibingkang20, yihongzhu23\}@mails.ucas.ac.cn

    xujg@ucas.ac.cn
}

\usepackage{bibentry}

\begin{document}

\maketitle

\begin{abstract}
Unsupervised sentence embedding representation has become a hot research topic in natural language processing. As a tensor, sentence embedding  has two critical properties: direction and norm.
Existing works have been limited to constraining only the orientation of the samples' representations while ignoring the features of their module lengths. To address this issue, we propose a new training objective that optimizes the training of unsupervised contrastive learning by constraining the module length features between positive samples.
We combine the training objective of \textbf{T}ensor's \textbf{N}orm \textbf{C}onstraints with ensemble learning to propose a new \textbf{S}entence \textbf{E}mbedding representation framework, \textbf{TNCSE}.
We evaluate seven semantic text similarity tasks, and the results show that TNCSE and derived models are the current state-of-the-art approach; in addition, we conduct extensive zero-shot evaluations, and the results show that TNCSE outperforms other baselines.


\end{abstract}

\begin{links}
\link{Code}{https://github.com/tianyuzong/TNCSE}
\end{links}

%

\section{Introduction}

Unsupervised sentence embedding representations have always been a research focus in natural language processing. In recent years, with the proposal of excellent pre-trained language models such as BERT\cite{BERT}, RoBERTa\cite{roberta}, Sentence-BERT\cite{sbert}, and so on, many works have been done to optimize further the quality of sentence embedding representations based on them.
Sentence embedding for unsupervised contrastive learning represented by SimCSE\cite{gao2021simcse} has attracted extensive research interest, which uses InfoNCE\cite{infonce} as the loss function, the dropout function of BERT-like models as the positive sample generation method, other samples in the batch as the soft-negative samples, and cosine similarity as the semantic similarity metric function for the unsupervised training; on this basis, ESimCSE\cite{esimcse} introduces near-antonym data augmentation, InfoCSE\cite{infocse} introduces auxiliary networks and adds masked language models, ArcCSE\cite{arccse} introduces margin for contrastive learning, EDFSE\cite{zong} employs the data augmentation strategy for multilingual round-trip translation(RTT) and ensemble learning, and RankCSE\cite{rankcse}, the current SOTA method, employs knowledge distillation of naive BERT-like models with existing checkpoints.

Although the above approaches have demonstrated in different directions that fine-tuning based on InfoNCE can further optimize SimCSE, they have all neglected an essential drawback of an InfoNCE with cosine similarity as a metric, i.e., cosine similarity can only measure the angle between the embedding representations of a pair of sentences and ignores the critical attribute of module of the embedding representations.
In fact, if the embedding representations of two sentences have similar angles in the high-dimensional space, but the difference in the module is significant, their semantics may still be very different, and the cosine similarity only cannot differentiate the semantic difference between these two sentences. Based on this intuitionistic phenomenon, it is necessary to introduce the difference in the module between the representations to optimize SimCSE further.

Meanwhile, ensemble learning has been shown to improve sentence embedding representations' quality significantly. For example, EDFSE uses six SimCSE-BERTs that have been fine-tuned by multilingual RTT data augmentation as submodels and directly sums the outputs of the six submodels to obtain an ensemble model, but such a large union brings a huge inference overhead; RankCSE, on the other hand, uses the ensemble learning of two existing checkpoints as ground-truth to distill the knowledge of a naive BERT-like model. 
So, is it possible to introduce norm constraints on sentence embedding representations into ensemble learning, and the model can train autonomously without relying on other checkpoints of the same type?

The answer is yes, and in this paper, we propose a new and intuitive training objective: for a pair of positive samples of the embedding tensor, the cosine similarity between them should be high, and both norms should be similar. In other words, the alignment between positive samples should constrain the embedding's angle and modulus simultaneously.
In terms of model structure, the multi-encoder structure can model richer sample features from multiple spaces compared to a single encoder. Considering the inference cost, we ensemble two encoders that have been unsupervised fine-tuned. 
For the semantic representation of a sample, we perform self-supervised cosine constraints within each encoder and implement mutually supervised norm constraints between the two encoders to achieve our proposed training objective.


Then, we evaluate our proposed method on seven semantic text similarity tasks. The experimental results show that TNCSE, added by norm constraints, outperforms EDFSE on BERT-base, which is three times as large as TNCSE and becomes the new state-of-the-art method. For a fair comparison, we also do ensemble learning on other baselines and comparisons, and the results show that TNCSE is still the best.
In addition, we evaluate TNCSE in a wider zero-shot on the MTEB\cite{mteb} list, and the results show that TNCSE also outperforms the baseline. We conduct a series of ablation experiments to evaluate the role of each part of the model, which illustrates the mechanism of TNCSE's action intuitively.
We summarize the main contributions of our work as follows:
\begin{itemize}
\item We are the first to intuitively propose a positive sample tensor modulus-constrained training objective based on unsupervised contrastive learning and demonstrate that it is effective.
\item We combine ensemble learning with a tensor norm constraint strategy to propose a new sentence embedding representation framework, TNCSE, and generalize a series of derived models.
\item We evaluate TNCSE and derived models on seven STS tasks, and the results show that TNCSE has become a new SOTA method for STS tasks. In addition, we have done a series of zero-shot evaluations, such as multilingual and cross-linguistic STS tasks, sentence classification, retrieval, and reranking, and the results show that TNCSE outperforms the baselines.
\end{itemize}

\section{Related Work}

InfoNCE(Noise Contrastive Estimation Loss\cite{infonce}) is a loss function for self-supervised learning, usually used to learn feature representations, which is based on the idea of information theory and learns the model parameters by comparing the similarity (usually cosine similarity) between positive and negative samples.
There have been some significant works combining InfoNCE loss and pre-trained language models BERT\cite{BERT}, RoBERTa\cite{roberta}, etc. to achieve training of sentence embedding representations for unsupervised contrastive learning.
Both SimCSE\cite{gao2021simcse} and ConSERT\cite{consert} use the idea of dropout to generate positive samples, and they have in common the use of cosine similarity as the only metric to discriminate between positive and negative samples. 
Subsequent work further optimizes the training of unsupervised sentence embeddings based on SimCSE, e.g., ESimCSE\cite{esimcse} provides the idea of proxemic data augmentation while associating the idea of MoCo's\cite{moco} Momentum Queue to improve the quality of SimCSE's representations; DiffCSE\cite{diffcse} introduces additional discriminator knowledge to assist the pre-trained models to mask linguistic modeling.
SNCSE\cite{SNCSE} employs a contrastive learning strategy that combines soft and negative samples with a bi-directional margin loss; and InfoCSE\cite{infocse} introduces an additional network for constructing a masked language model;
EDFSE\cite{zong} constructs huge ensemble models by training multiple encoders through RTT data augmentation; the current SOTA approach, RankCSE\cite{rankcse}, uses dual-teacher ensemble learning and distillation to train encoder.
The commonality among these existing works is that they all introduce the InfoNCE loss, which employs the cosine angle between embeddings of individual samples to determine similarity. The crucial factor of the norm of the embedding tensor has been overlooked, so we propose an unsupervised ensemble learning framework with embedding representation norm constraints, to enhance the model's ability to discriminate the positive and negative samples.

\begin{figure}
\centering
\includegraphics[scale=0.7]{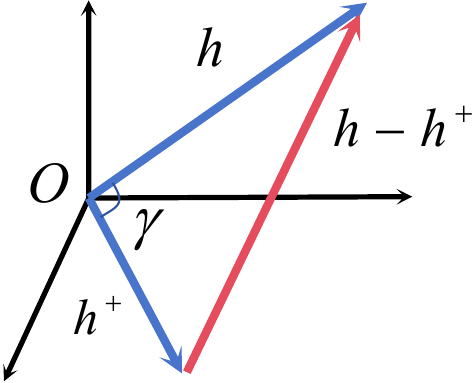}
\captionsetup{width=0.475\textwidth}
\caption{The figure denotes the subtraction of the semantic representation tensor in 3D space.}
\label{xyz}
\end{figure}

\begin{figure}
\centering
\includegraphics[scale=0.7]{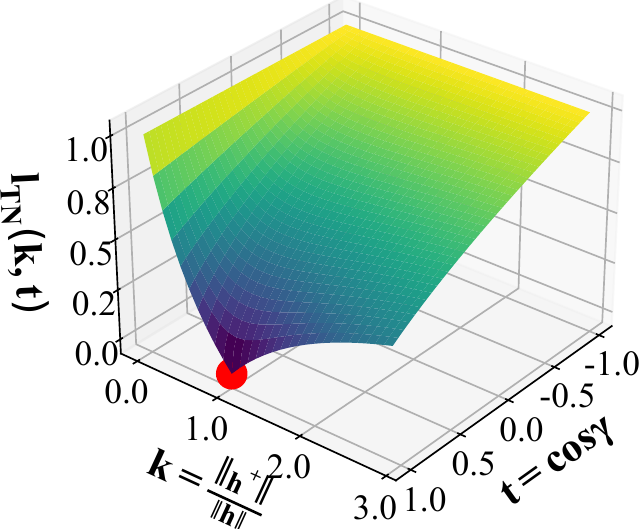}
\captionsetup{width=0.475\textwidth}
\caption{The figure denotes the binary function image of the constraint loss of the tensor norm for independent variables $k$ and $t$.}
\label{ltn0000}
\end{figure}

\section{Method}

In this section, we start by presenting the norm constraints imposed on semantic representation tensors. Subsequently, we incorporate ensemble learning strategies, leading to the elaboration of our proposed framework, the Tensor Norm Constrained Semantic Embedding (TNCSE).



\subsection{Norm Constraints on Representations}

Traditional unsupervised text embedding representation training typically uses the cosine similarity between representation tensors as the primary metric for determining the degree of similarity between sentence pairs, but this only considers the “direction” of the tensor properties and ignores the “magnitude”. \citet{alignment-uniformity} believes that semantic representations can be projected onto a hypersphere. The distribution of positive samples should be aligned as much as possible. The soft negative samples should be distributed uniformly on the hypersphere. However, to distinguish semantic similarity among many texts in a fine-grained way, we should consider both the direction and the magnitude of the representation tensor. For a semantic representation of a pair of positive samples, intuitively, the angle between them should be as slight as possible, and the magnitudes should be identical. Therefore, in our approach, we model the 2-paradigm(denoted hereafter as the \textbf{norm}) of the semantic representation tensor as its “magnitude”, to perform constraints, and propose a loss based on the norm constraints as shown in Eq. \ref{ltn0}:

\begin{equation}
    l_{TN}(h,h^{+} )= \frac{\left \| {h} -{h }^{+}   \right \| }{\left \| {h}  \right \|+\left \| {h}^{+}   \right \|  }.
    \label{ltn0}
\end{equation}

In Eq. \ref{ltn0}, $h$ and $h^{+}$ denote the embedding representations of a sample and its positive samples, $\left \|  \cdot \right \| $ denotes the norm of tensor, and we come to demonstrate the rationality of this loss. For visualization, we report this process in a 3-dimensional space. As shown in Fig. \ref{xyz}, the two semantic representation tensors $h$ and $h^{+}$ of positive samples have difference tensor $h-h^{+}$. 
These three tensors construct a triangle with sides $\left \| h \right \| $, $\left \| h^{+}  \right \| $, and $\left \|h- h^{+}  \right \| $,
and let the angle between $h$ and $h^{+}$ be $\gamma $. Then, according to the cosine theorem, we have:

\begin{equation}
    l_{TN}(h,h^{+} )= \frac{\sqrt{\left \| h  \right \|^{2}  +\left \| h ^{+}  \right \|^{2}  -2 \left \| h  \right \| \left \| h^{+}  \right \| \cos \gamma  } }{\left \| h  \right \|+\left \| h^{+}   \right \|  } .
    \label{costhe}
\end{equation}

Since the norm of semantic representation are all nonzero, we make $\left \| {h}^{+}   \right \| =k\cdot \left \| {h}  \right \|$, $t=\cos \gamma$, where $k\in \left ( 0,+\infty  \right ) $ and $t\in \left [ -1,1 \right ] $. Bringing above into the Eq. \ref{costhe}, there is:

\begin{equation}
    l_{TN}\left ( k,t \right )  =\frac{\sqrt{1+k^{2}-2\cdot kt } }{1+k}.
    \label{ltnkt}
\end{equation}

We plot the image of the binary function $l_{TN}$ for the independent variables $k$ and $t$ over part of the domain of definition, as shown in Fig. \ref{ltn0000}. $l_{TN}$ takes a minimal value when $k$ and $t$ are equal to 1, respectively, in other word, when $\cos \gamma =1$ and $\left \| {h}^{+}   \right \| = \left \| {h}  \right \|$. This result aligns with our intuition, our proposed $l_{TN}$ can simultaneously constrain the two metrics of cosine similarity and modulus length between positive samples of the semantic representation tensor.



\begin{figure*}
\begin{center}
    \includegraphics[scale=0.587]{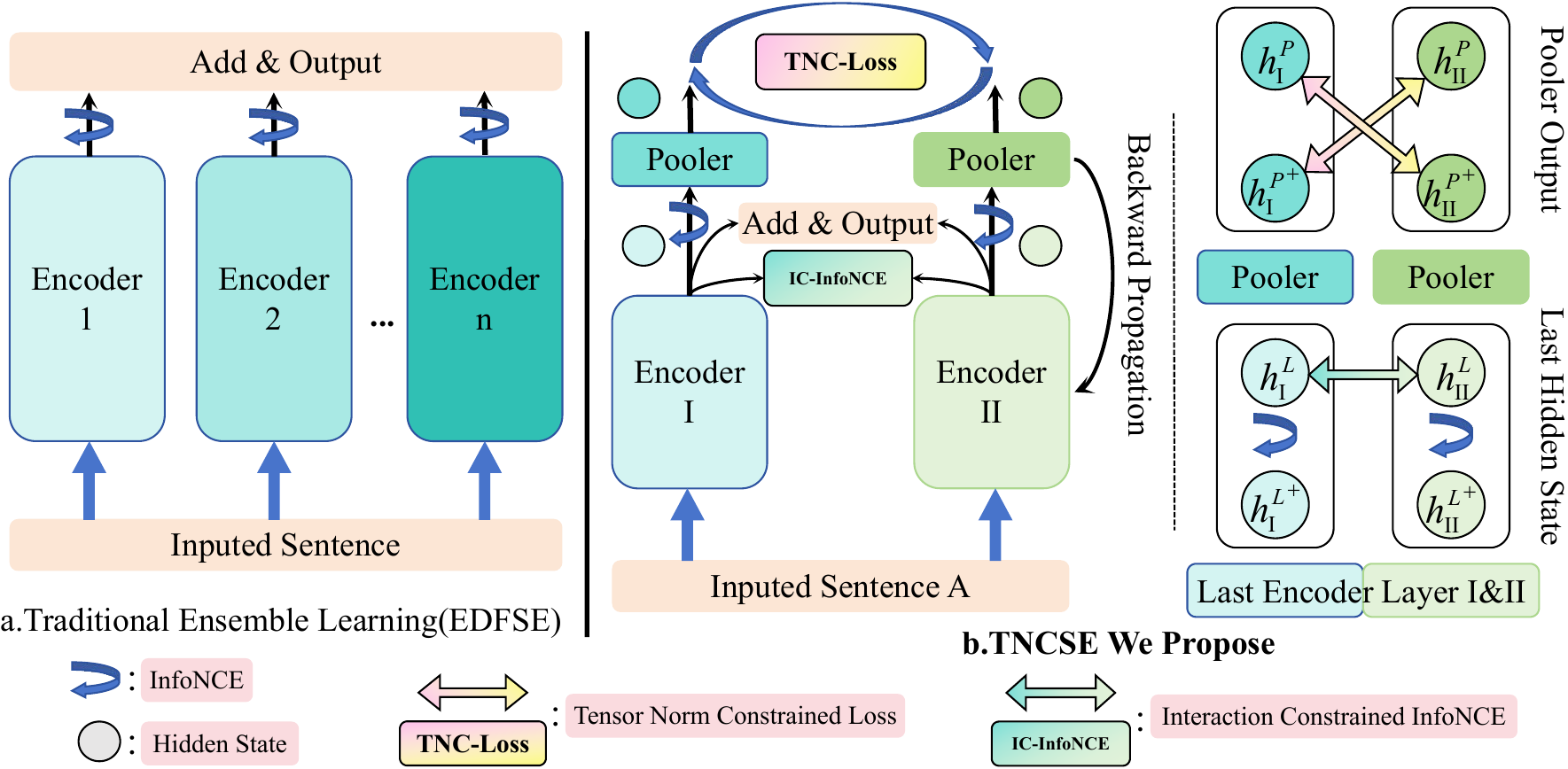}
\captionsetup{width=\textwidth} 
\caption{The left side of the solid line is the traditional ensemble learning method represented by EDFSE, and the right is our proposed new method based on semantic tensor with norm constraints.
When TNCSE is training, sample $A$ passes through two encoders simultaneously, obtains the last hidden state, does InfoNCE$(L_{NCE})$ and ICNCE$(L_{ICNCE})$, then passes through the corresponding pooler layer and does TNC$(L_{ICTN})$ with norm constraints crossly, respectively.
When TNCSE is inference, sample $A$ passes through two encoders simultaneously, and then the two last hidden states are directly summed up as the output.}
\label{mainfigure1}
\end{center}
\end{figure*}

\subsection{Model Structure Design}

This subsection describes our proposed sentence embedding representation framework TNCSE based on tensor norm constraints, as shown in Fig. \ref{mainfigure1}b. The traditional large-scale ensemble model EDFSE (Fig. \ref{mainfigure1}a) uses six encoders, each of which has been pretrained with multilingual RTT data augmentation and unsupervised SimCSE. This ensemble strategy endows the model with the advantage of diverging the embeddings from different encoders as much as possible, thereby increasing the "intrinsic rank" of the joint representation, leading to enhanced capability in discriminating between semantically similar sentences.
We follow EDFSE's data augmentation approach, but only employ two encoders that have been data augmented and fine-tuned by unsupervised SimCSE for reducing EDFSE's inference overhead.
For each of the two encoders of TNCSE-BERT and TNCSE-RoBERTa, we randomly chose four target languages for RTT Data Augmentation (EN-Target-EN).
Each of these encoders ends up with its pooler layer. We denote these two encoders as Encoder I and Encoder II, respectively.

When the TNCSE is training, due to the dropout in the encoder, each sentence is fed into Encoder I and Encoder II twice to obtain positive samples, respectively. The last hidden state of both encoders contains two similar embeddings, and these four representations are denoted as $h_{\mathrm {I}   }^{L} $, ${h_{\mathrm {I}   }^{L} }^{+} $, $h_{\mathrm {II}   }^{L} $, and ${h_{\mathrm {II}   }^{L} }^{+} $, respectively. Since both are representations of the same sentence sample, two of these four hidden states are positive samples of each other, so we set the respective InfoNCE and the \textbf{i}nteraction-\textbf{c}onstrained InfoNCE(ICNCE), InfoNCE is denoted as Eq. \ref{infon}:

\begin{equation}
   L_{N C E}\left(h_{i}, h_{i}^{+}\right)=-\log \frac{e^{ \frac{\operatorname{sim}\left(h_{i}, h_{i}^{+}\right)}{\tau } }}{e^{ \frac{\operatorname{sim}\left(h_{i}, h_{i}^{+}\right)}{\tau } }+\sum e^{ \frac{\operatorname{sim}\left(h_{i}, h_{j}^{-}\right)}{\tau } }} .
    \label{infon}
\end{equation}
Here, $i\in \left \{ \mathrm {I},\mathrm {II}       \right \} $ indicates from Encoder I or Encoder II. $h_{i}$ and $h_{i}^{+}$ are mutually positive samples, and $h_{j}^{-}$ is the other samples within the current $batch\ size$ as soft-negative samples. We follow the SimCSE setting with $ \tau = 0.05$; $sim(\cdot)$ is the cosine similarity. ICNCE is then denoted as Eq. \ref{ICNCE}:

\begin{equation}
    L_{ICNCE} =L_{NCE}\left ( h_{\mathrm {I}  }^{L} ,h_{\mathrm {II}   } ^{L}  \right ).
    \label{ICNCE}
\end{equation}
InfoNCE serves to continue training and maintain the general direction of the encoder's gradient update, while ICNCE is designed to precondition subsequent tensor norm loss.

Due to the norms of the last hidden states in BERT-like models converging to a uniform value, we cannot effectively apply tensor norm constraints to the last hidden states of positive sample pairs. However, we notice that the last pooler layer in a BERT-like model is a single feedforward neural network, which can be endowed with normative features by projecting the last hidden state onto a space of the same dimension. Accordingly, we naturally employ the output of the pooler layer for tensor norm constraints.
Moreover, we find that the cosine similarity of the last hidden state of the positive sample pairs can be used as a nondeterministic prompt for the $l_{TN}$. 
Thus, we amend the $l_{TN}$ to be Eq. \ref{ltn}:

\begin{equation}
    L_{TN}\left ( h_{i},h_{j}^{+}  \right ) =-\log \left ( sim \left ( h_{\mathrm {I}}^{L},{h_{\mathrm {II}}^{L}}   \right )  \right )  \frac{\left \| h_{i}^{P}-h_{j }^{P^{+} }    \right \| }{\left \| h_{i}^{P}  \right \| +\left \| h_{j }^{P^{+} }   \right \| } ,
    \label{ltn}
\end{equation}
where $i,j\in \left \{ \mathrm {I},\mathrm {II}       \right \} $ and $i\ne j$. $h_{\mathrm {I}   }^{P} $, ${h_{\mathrm {I}   }^{P} }^{+} $, $h_{\mathrm {II}   }^{P} $, and ${h_{\mathrm {II}   }^{P} }^{+} $ denote the pooler output of these hidden states, respectively. Here we set up an \textbf{i}nteraction \textbf{c}onstraint on the \textbf{t}ensor \textbf{n}orm(ICTN), denoted as Eq. \ref{ICTN}:

\begin{equation}
    L_{ICTN}=L_{TN}\left ( h_{\mathrm {I}}  ,h_{\mathrm {II}    }^{+} \right ) +L_{TN}\left ( h_{\mathrm {II}}  ,h_{\mathrm {I}    }^{+} \right ) 
    \label{ICTN}.
\end{equation}

The purpose of designing $L_{ICTN}$ in this way is to enhance the norm alignment of positive samples in the joint representation space of the two encoders. Finally, the loss of the TNCSE is denoted as Eq. \ref{allloss}:

\begin{equation}
    L=\sum_{i\in \left \{ \mathrm {I},\mathrm {II}       \right \} }L_{NCE}\left ( h_{i}^{L},{h_{i}^{L}}^{+}   \right ) +L_{ICNCE}+L_{ICTN} .
    \label{allloss}
\end{equation}

We have demonstrated through ablation experiments that each of the above components is necessary. When we are testing TNCSE, for an input sentence, we use two encoders to encode the sentence separately, and the obtained last hidden state is directly summed up without going through the pooler layer as the output of TNCSE.

\begin{table*}
\begin{center}
\label{bigresult}
\begin{tabular}{lcccccccc}
\toprule[2pt]
\textbf{Model}           & \textbf{STS12} & \textbf{STS13} & \textbf{STS14} & \textbf{STS15} & \textbf{STS16} & \textbf{STSB} & \textbf{SICKR} & \textbf{Avg.}  \\ \hline
\multicolumn{9}{c}{\textbf{BERT-base}  }                                                       \\ \hline
SimCSE\cite{gao2021simcse}$\star$           & 68.40 & 82.41 & 74.38 & 80.91 & 78.56 & 76.85 & 72.23  & 76.25 \\
DiffCSE\cite{diffcse}$\star$          & 72.28 & 84.43 & 76.47 & 83.90 & 80.54 & 80.59 & 71.23  & 78.49 \\
ESimCSE\cite{esimcse}$\star$          & 73.40 & 83.27 & 77.25 & 82.66 & 78.81 & 80.17 & 72.30  & 78.27 \\
ArcCSE\cite{arccse}$\star$           & 72.08 & 84.27 & 76.25 & 82.32 & 79.54 & 79.92 & 72.39  & 78.11 \\
InfoCSE\cite{infocse}$\star$          & 70.53 & 84.59 & 76.40 & \textbf{85.10} & \textbf{81.95} & \textbf{82.00} & 71.37  & 78.85 \\
PromptBERT\cite{PromptBERT}$\star$                & 71.56 & 84.58 & 76.98 & 84.47 & 80.60 & 81.60 & 69.87  & 78.54\\
SNCSE\cite{SNCSE}$\star$                & 70.67 & 84.79 & 76.99 & 83.69 & 80.51 & 81.35 & \textbf{74.77}  & 78.97\\
WhitenedCSE\cite{whencse}$\star$        &74.03 &\textbf{84.90} &76.40 &83.40 &80.23 &81.14 &71.33 &78.78 \\ \hdashline
EDFSE\cite{zong}$\star$            & 74.48          & 83.14          & 76.39          & 84.45 & 80.02          & 81.97 & 72.83           & 79.04 \\
\textbf{TNCSE}   & \textbf{75.52} & 83.91 & \textbf{77.57} & 84.97 & 80.42 & 81.72 & 72.97  &  \textbf{79.58} \\
EDFSE D\cite{zong}$\star$    & 74.50 & 83.61          & 76.24          & 84.02          & 80.44          & 81.94          & 74.16  & 79.27  \\ 
\textbf{TNCSE D}   & \textbf{75.42}  & 84.64 & \textbf{77.62} & 74.92 & 80.50 & 81.79 & 73.52  & \textbf{79.77} \\ \hline

\cellcolor{blue!20} RankCSE\cite{rankcse}$\clubsuit  $   & 74.61  & 85.70 & 78.09 & 84.64 & 81.36 & 81.82 & 74.51  & 80.10 \\ \hdashline
\cellcolor{blue!20} RankCSE+UC   & 73.29  & \textbf{85.90} & 78.16 & 85.90 & \textbf{82.52} & 83.13 & \textbf{73.36}  & 80.32 \\ 
\cellcolor{blue!20} \textbf{TNCSE+UC }           & \textbf{75.79}  & 85.27 & \textbf{78.67} & \textbf{85.99} & 82.01 & \textbf{83.16} & 73.01  & \textbf{80.56} \\ \hdashline
\cellcolor{blue!20} RankCSE+UC \textbf{D}   & 72.99  & \textbf{85.72} & 77.73 & 84.93 & \textbf{81.86} & 82.43 & \textbf{74.35}  & 80.00 \\ 
\cellcolor{blue!20} \textbf{TNCSE+UC D} & \textbf{75.95}  & 85.31 & \textbf{78.50} & \textbf{85.69} & 81.86 & \textbf{83.03} & 73.89  &  \textbf{80.60} \\ \toprule[1.5pt]
 
\multicolumn{9}{c}{\textbf{RoBERTa-base} }                                                     \\ \toprule[1.5pt]
SimCSE\cite{gao2021simcse}$\star$           & 70.16 & 81.77 & 73.24 & 81.36 & 80.65 & 80.22 & 68.56  & 76.57 \\
DiffCSE\cite{diffcse}$\star$          & 70.05 & 83.43 & 75.49 & 82.81 & 82.12 & 82.38 & 71.19  & 78.21 \\
ESimCSE\cite{esimcse}$\star$          & 69.90 & 82.50 & 74.68 & 83.19 & 80.30 & 80.99 & 70.54  & 77.44 \\
PromptBERT\citeyearpar{PromptBERT}$\star$                & 73.94 & \textbf{84.74} & \textbf{77.28} & \textbf{84.99} & 81.74 & 81.88 & 69.50  & 79.15\\
SNCSE\cite{SNCSE}$\star$                & 70.62 & 84.42 & 77.24 & 84.85 & 81.49 & \textbf{83.07} & 72.92  & 79.23\\
WhitenedCSE\cite{whencse}$\star$   &70.73 &83.77 &75.56 &81.85 &\textbf{83.25} &81.43 &70.96 &78.22 \\ 
IS-CSE\cite{iscse}$\star$           & 71.39 & 82.58 & 74.36 & 82.75 & 81.61 & 81.40 & 69.99  & 77.73 \\ \hdashline

EDFSE\cite{zong}  & 72.67 & 83.00 & 75.69 & 84.07 & 82.01 & 82.53 & 71.92  & 78.84 \\         
\textbf{TNCSE}    & 74.11 & 84.00          & 76.06          & 84.80 & 81.61          & 82.68          & 73.47           & \textbf{79.53} \\  
EDFSE D\cite{zong}           & 71.04          & 81.08          & 77.04          & 83.08          & 81.96          & 82.36          & \textbf{74.54}  & 78.73 \\ 
\textbf{TNCSE D}     & \textbf{74.56} & \textbf{84.74} & 76.30 & 84.89 & 81.70 & 83.01 & 74.18  &\textbf{79.91} \\
\hline
\cellcolor{blue!20} RankCSE\cite{rankcse}$\clubsuit  $        & 69.09          & 81.15 & 73.62 & 81.31          & 81.43          & 81.22          & 70.08           & 76.84  \\ \hdashline
\cellcolor{blue!20} RankCSE+UC     & 74.18 & 84.06 & \textbf{77.72} & 83.26 & 79.81 & 81.25 & 72.58  & 78.98 \\
\cellcolor{blue!20} \textbf{TNCSE+UC}     & \textbf{74.52} & \textbf{85.26} & 77.63 & \textbf{85.85} & \textbf{82.62} & \textbf{83.65} & \textbf{73.35}  & \textbf{80.41} \\ 
\hdashline
\cellcolor{blue!20} RankCSE+UC \textbf{D}     & 68.55 & 82.23 & 73.61 & 81.28 & 81.28 & 80.98 & 71.01  & 76.99 \\
\cellcolor{blue!20} \textbf{TNCSE+UC D}     & \textbf{74.14} & \textbf{83.86} & \textbf{76.09} & \textbf{84.07} & \textbf{81.59} & \textbf{82.90} & \textbf{73.55}  & \textbf{79.46} \\
\toprule[2pt]
\end{tabular}
\caption{ This table reports the results of the TNCSE and baseline evaluation on the seven STS tasks. $\star $ denotes results derived from the original paper. Since RankCSE\cite{rankcse} has not officially open-sourced any code or checkpoints, $\clubsuit $ denotes the result of a third-party open-source code replication\protect\footnotemark. \textbf{D} denotes distillation to a single encoder. 
}
\label{stsresult}
\end{center}
\end{table*}
\footnotetext{\url{https://github.com/perceptiveshawty/RankCSE}}

\begin{table*}
\setlength\tabcolsep{4.8pt} 
\begin{center}
\begin{tabular}{lccccccccc}
\toprule[1pt]
\textbf{Tasks(Avg.)}          & \textbf{SimCSE} & \textbf{ESimCSE} & \textbf{DiffCSE} & \textbf{InfoCSE} & \textbf{SNCSE} & \textbf{RankCSE} & \textbf{TNCSE} & \textbf{+D} & \textbf{+UC D} \\ 
\toprule[0.75pt]
\textbf{STS17(11 Acc)}  & 34.22          & 35.64             & 31.61             & 35.95             & 22.64           & 37.59             & 36.32  &  35.41    & \textbf{37.73}           \\
\textbf{STS22(18 Acc)}  & 32.78          & 36.75             & 34.33             & 26.48             & 23.61           & 37.01             & \textbf{39.34}  &  \textbf{38.95}    & \textbf{39.26}           \\
\textbf{Classification(30 Acc)}  & 58.15           & 58.26            & 57.44            & 58.15            & 58.05             & 58.15            & \textbf{58.38} & \textbf{58.59}   & \textbf{58.84}     \\ 
\textbf{Reranking(9 Map)}      &   36.78 &	38.12 &	37.13 &	38.36 &	31.95  &	37.11 	& 38.15  & \textbf{38.93}  &	38.29 \\
\textbf{Retrieval(30 Map@10)}      & 18.98           & 21.13            & 19.09            & 21.54            & 16.88            & 18.33            & \textbf{22.10} & \textbf{23.16}      & \textbf{22.14}   \\ 
\toprule[1pt]

\end{tabular}
\caption{This table reports the results of the zero-shot evaluation of TNCSE and baseline on MTEB, with the number of tasks and metric indicated in parentheses.
Details of the results are reported in Appendix IV.}
\label{mtebtable}
\end{center}
\end{table*}

\begin{table*}
\begin{center}
    
\begin{tabular}{lccccccc}
\toprule[1pt]
\textbf{Model(Dual Encoder)} & \textbf{SimCSE} & \textbf{ESimCSE} & \textbf{DiffCSE} & \textbf{InfoCSE} & \textbf{RankCSE} & \textbf{TNCSE(Untrained)} & \textbf{TNCSE} \\ 
\toprule[0.75pt]
\textbf{7 STS Avg.}          & 77.97      &      78.51            &      77.89            &   77.05               &  78.22                &    78.27               & \textbf{79.58}                \\ 
\toprule[1pt]
\end{tabular}
\caption{We reports the results of ensemble learning done by a set of encoders obtained by adding unlabelled SICKR dataset and the 2 RTT augmented datasets to which each baseline is trained(Default hyperparameters) and compares them with TNCSE and the two encoders untrained by TNCSE. The inconsistency between the dual SimCSE and TNCSE-Untrained results is due to the fine-tuning of the learning rate we have done during pre-training, but the dependence on external knowledge is the same.}
\label{baselinedoensemble}
\end{center}
\end{table*}

\section{Experiment}

\subsubsection{Setup}
In TNCSE, we first employ two BERT-bases or RoBERTa-bases for pre-training with Google Translate for two different RTT Data Augmentation and unsupervised SimCSE, respectively. The dataset employed is the 1M Wiki corpus\footnote{\url{https://huggingface.co/datasets/princeton-nlp/datasets-for-simcse}} and unlabelled data from SICKR\cite{sickr}. Hyperparameter settings and target languages of RTT are reported in Appendix I. We follow the SimCSE setup, with Spearman's correlation of the model on the STS-B validation set as the checkpoint saved metric. We follow SimCSE with CLS as the pooling method and report the impact of pooling in Appendix II.


\subsubsection{Tasks}
We evaluate TNCSE and baseline on 7 STS tasks with the SentEval\cite{senteval} tool, including STS12-16\cite{sts12,13,14,15,16}, STS-B\cite{stsb}, and SICKR; we then conduct extensive zero-shot tests on the sentence-embedding task list MTEB\cite{mteb}, which includes both multilingual and cross-language STS tasks, such as STS17\cite{sts17}, STS22\cite{sts22}, and randomly select 30 sentence classification, 30 retrieval and all reranking tasks available for the test set.


\subsubsection{Experimental Results}

We report the experimental results for the seven STS tasks in Table \ref{stsresult}, where TNCSE outperforms the previous baselines. Since TNCSE is an ensemble model containing two encoders, we follow EDFSE's distillation approach and distill the knowledge from TNCSE to a naive BERT/RoBERTa, and obtain TNCSE-D, which also outperforms baselines as a single-encoder model.
Compared to a traditional ensemble model like EDFSE, we propose an efficient training objective that makes the inference overhead of TNCSE only one-third of EDFSE-BERT, but with better results on the 7 STS task.

Since RankCSE relies on the knowledge of the existing powerful checkpoints, SimCSE-base and SimCSE-large, to train the encoder in a knowledge distillation approach, and RankCSE is not trained independently, we consider that it is inappropriate to compare RankCSE with TNCSE and other baselines directly. We ensemble and distill RankCSE and TNCSE with unsupervised checkpoint InfoCSE in the same way, respectively, and get the ensemble model (UC) and distillation model (UC D) to evaluate; in other words, we prove that TNCSE is better than RankCSE by proving that TNCSE+InfoCSE is better than RankCSE+InfoCSE. In summary, TNCSE and derived models are the current SOTA methods for unsupervised STS tasks.
Due to computational resource constraints and the unavailability of some test sets, we randomly select 30 sentence classification tasks and 30 retrieval tasks, 29 multilingual and cross-linguistic STS tasks (STS17, STS22), and all 9 reranking tasks for which test sets are available on the MTEB, and report the experimental results in Table \ref{mtebtable}, where the overall results show that the TNCSE and derived models outperform the baseline approach.

\subsubsection{Significance Test}

\begin{figure}
\centering
\includegraphics[scale=0.5]{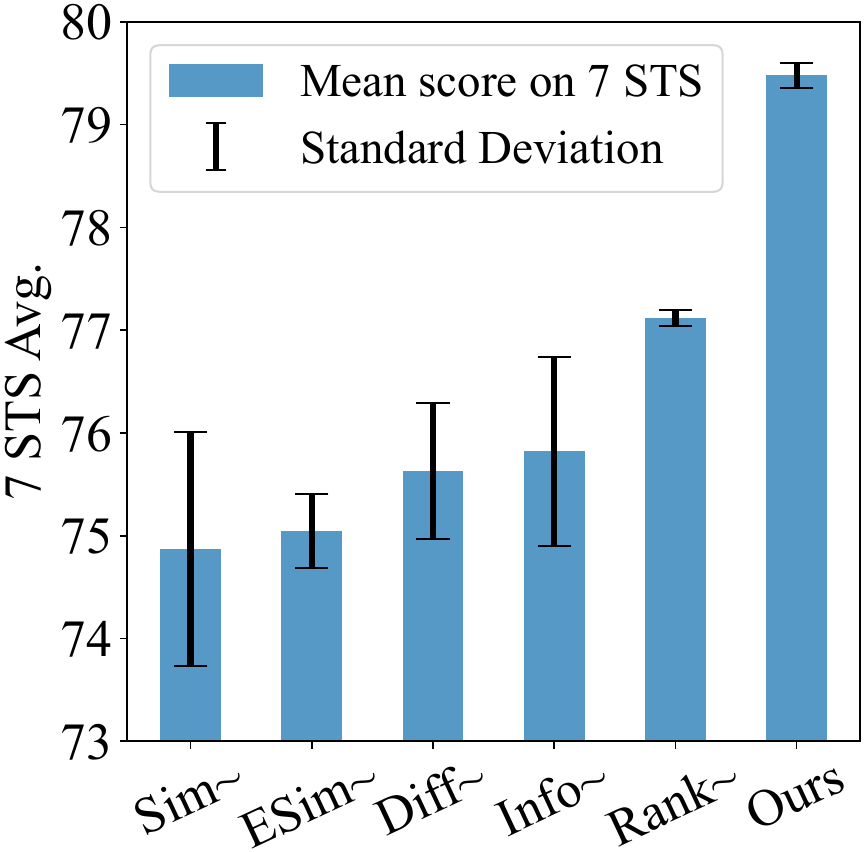}
\captionsetup{width=0.475\textwidth}
\caption{The figure reports the results of the significance test. We specify the random seeds are 1 to 5, other hyperparameters are defaulted, training set is uniformly Wiki1M and unlabelled SICKR.}
\label{significationdpf}
\end{figure}

Since the BERT-like model is single encoder and training is susceptible to random seeds, we design TNCSE with dual encoder structure, which is more stable than single encoder. We report the significance test of TNCSE and baselines in Figure \ref{significationdpf}.




\begin{figure}
\centering
\includegraphics[scale=0.375]{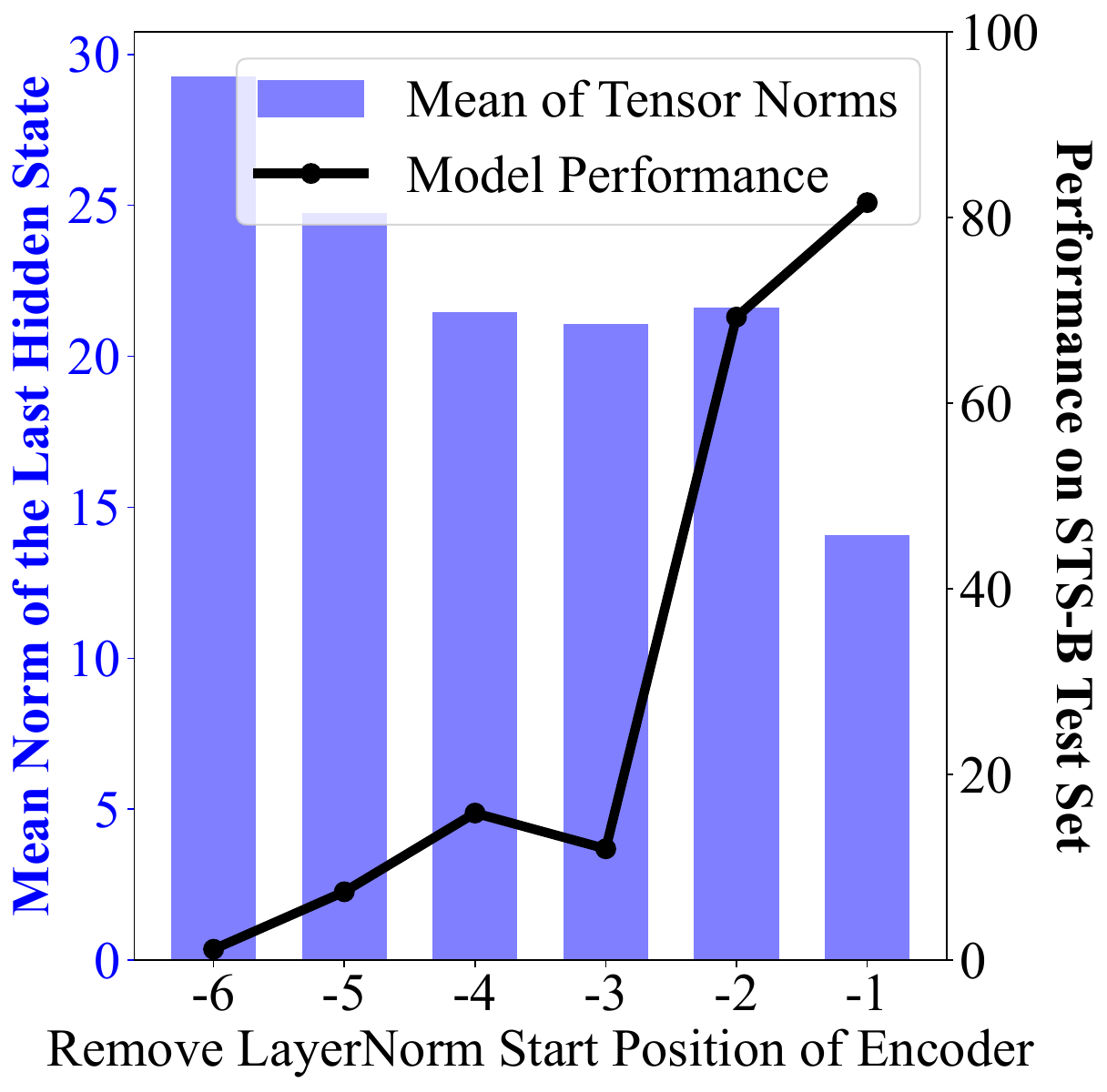}
\captionsetup{width=0.475\textwidth}
\caption{The bar and line graphs separately represent the norm mean of the Last hidden state and the model's performance on the STS-B validation set for different LayerNorm settings.}
\label{discusspdf}
\end{figure}

\section{Ablation Studies and Discussion}

\subsection{Alignment and Uniformity}
\begin{figure}
\centering
\includegraphics[scale=0.425]{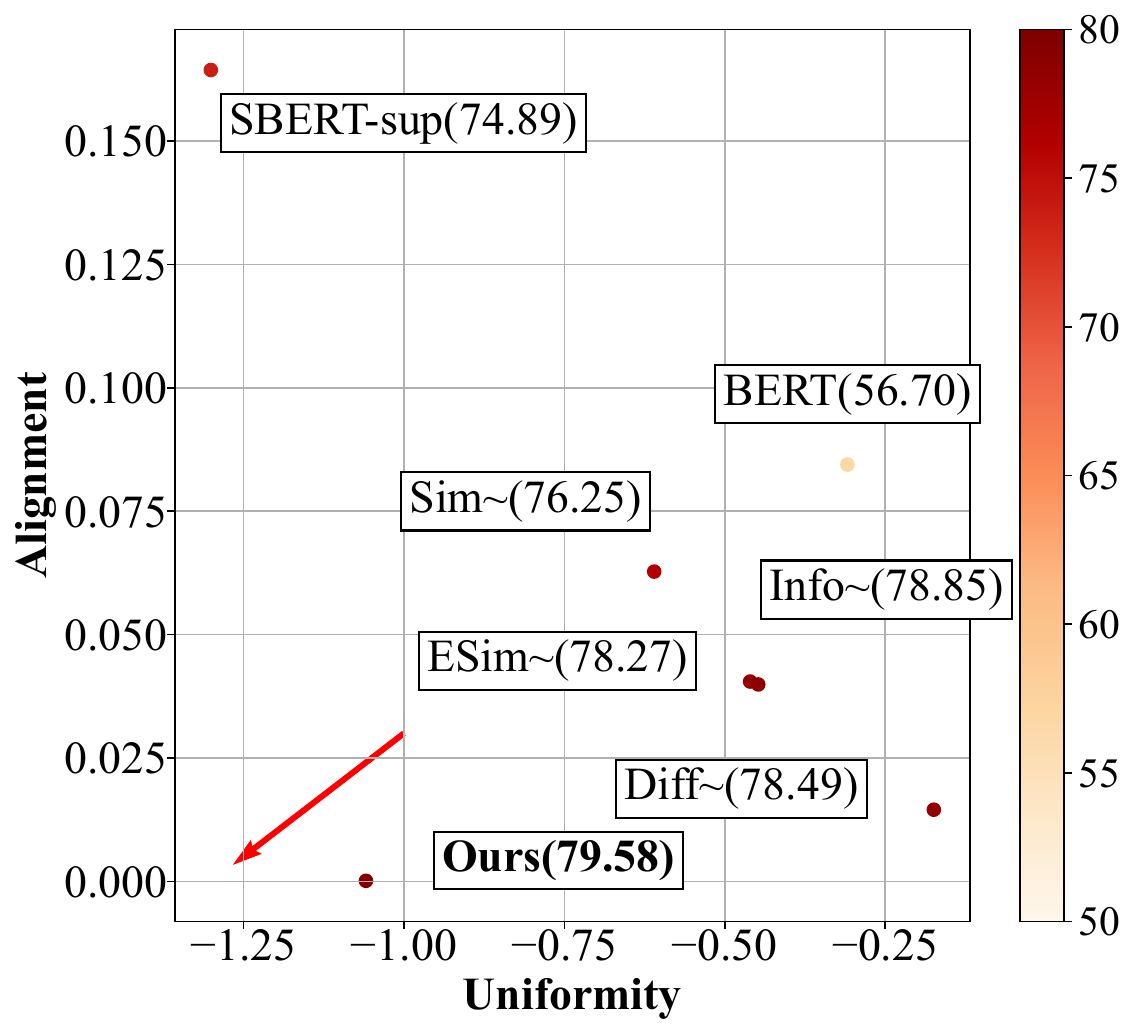}
\captionsetup{width=0.475\textwidth}
\caption{The alignment and uniformity metrics for TNCSE and baselines; the better the distribution is down the left. Darker the circle color, better the model performs on 7 STS.}
\label{alignpdf}
\end{figure}

\citet{alignment-uniformity} proposes that two critical metrics for measuring sentence embedding representations are uniformity and alignment, respectively, which we have already introduced in the model section, both of which are as small as possible. We report the scores of these two metrics for TNCSE and baselines in Figure \ref{alignpdf}. Due to the introduction of the tensor-norm alignment strategy in TNCSE, the loss of TNCSE concerning conventional angle alignment is superior. Because TNCSE unites ensemble learning, uniformity is also superior. We report on the detailed formulation of alignment and uniformity in Appendix III.

\begin{table}
\begin{center}
\begin{tabular}{lc}
\toprule[1pt]
\textbf{Loss Choice}         & \multicolumn{1}{l}{\textbf{7 STS Avg.}} \\ 
\toprule[0.75pt]
\textbf{None}(Dual Encoder Untrained)                  &78.27  \\
\textbf{$L_{NCE}$}                 & 78.42                                   \\
\textbf{$L_{ICNCE}$}               & 78.71                                   \\
\textbf{$L_{ICTN}$}                 & 78.38                                   \\
\textbf{$L_{NCE}+L_{ICNCE}$}           & 78.71                                   \\
\textbf{$L_{NCE}+L_{ICTN}$}             & 79.53                                   \\
\textbf{$L_{ICNCE}+L_{ICTN}$}           & 79.43                                   \\
\textbf{$L_{NCE}+L_{ICNCE}+L_{ICTN}$(Ours)} & \textbf{79.58}                          \\ 
\toprule[1pt]
\end{tabular}
\captionsetup{width=0.45\textwidth} 
\caption{This table combines each of the loss functions to explore the contribution of each to model training. \textbf{None} denotes a direct ensemble of two SimCSE-trained encoders. All experiments use CLS pooling method.}
\label{Loss Choice}
\end{center}
\end{table}

\subsection{Ablation of The Loss Function}

Since our loss function contains three terms, InfoNCE, ICNCE, and ICTN, this subsection investigates the performance gain from each term and analyses the reasons. The results of the ablation experiments are reported in Table \ref{Loss Choice}.

\subsubsection{$L_{NCE}$ Only}
Since TNCSE employs a dual encoder structure, the sentence embedding is modeled through ensemble learning. If only unsupervised training of InfoNCE is performed for each encoder, which does not directly improve the ensemble learning, and result is improved insignificantly.

\begin{table}
\begin{center}
\begin{tabular}{lll}
\toprule[1pt]
\textbf{Model} & \textbf{SimCSE}         & \textbf{TNCSE-Single}              \\ 
\toprule[0.75pt]
\textbf{STS-B testset} & \multicolumn{1}{c}{77.75} & \multicolumn{1}{c}{\textbf{78.37}} \\ 
\toprule[1pt]
\end{tabular}
\captionsetup{width=0.45\textwidth} 
\caption{This table compares the training results of SimCSE and TNCSE with only a single encoder under the same training set.}
\label{tncsesingleencoder}
\end{center}
\end{table}

\subsubsection{$L_{ICNCE}$ Only}
ICNCE is a boosted version of InfoNCE. ICNCE optimizes the tensor angle of positive and negative samples between two encoders, which works in ensemble learning. However, since this does not introduce a new training objective, it still belongs to the incremental training of InfoNCE, so the improvement is limited.

\subsubsection{$L_{ICTN}$ Only}
Since $L_{ICTN}$ (Eq. \ref{ICTN}) accomplishes norm constraint on the output of the encoder's pooler layer, but in the pre-training of the encoder, only the last hidden state is taken out to do the self-supervised training, and it does not go through the pooler layer, there is a margin between the last hidden state and the pooler output.
We will analyze in the discussion subsection that the last hidden state cannot directly use norm loss because it lacks norm features, so using only the pooler output to constrain the tensor norm in ensemble learning is almost ineffective with the CLS pooling method.

\subsubsection{$L_{NCE}+L_{ICNCE}$}
Under the joint constraints of $L_{NCE}$ and $L_{ICNCE}$, the two encoders perform self-supervised continuation training and, at the same time, mutually supervise each other's hidden states, which facilitates fine-grained representations of samples by ensemble learning, but still falls under the tensor's angle constraints. Therefore, the improvement is minor.

\subsubsection{$L_{NCE}+L_{ICTN}$ and $L_{ICNCE}+L_{ICTN}$}
The combination of the tensor's norm constraints and either of the angle constraints resulted in a significant improvement in training. This is due to the two properties of the trained tensor, allowing the two encoders to more clearly distinguish between two soft negative samples with similar representations after ensemble learning, improving the model performance.

\subsection{Ablation of Ensemble Learning}
Since TNCSE is dual-encoder structured, although we have distilled its knowledge to a single encoder, we still ablate its structure. We keep only one naive encoder, BERT-base, whose pooler layer outputs a pair of positive samples to accomplish the constraints of $L_{TN}$ (Eq. \ref{ltn}) and the last hidden state to accomplish the constraints of self-supervised InfoNCE. In other words, this structure only increases the $L_{TN}$ loss relative to Unsup-SimCSE, and we retrained unsupervised SimCSE in the same experimental setting with the same training set for both Wiki1M and unlabelled SICKR. We report the model's performance on the STS-B test set in Table \ref{tncsesingleencoder}. Compared with the Unsup-SimCSE, which includes the same baseline knowledge, the model with the added $L_{TN}$ loss is more effective than the unsupervised SimCSE.
In addition, we put several representative baselines through the same two different RTT pre-training to get a dual encoder, and then compare it with TNCSE. The results are reported in Table \ref{baselinedoensemble}.

\begin{table}
\begin{center}
\begin{tabular}{lcc}
\hline
\textbf{Datasets} & \multicolumn{1}{l}{\textbf{Wiki1M}} & \multicolumn{1}{l}{\textbf{+Unlabelled SICKR}} \\ \hline
\textbf{SimCSE}   &  74.3                                   &  75.9(+1.6)                                              \\
\textbf{ESimCSE}  &    75.8                                 &    76.7(+0.9)                                            \\
\textbf{DiffCSE}  &  75.2                                   &  78.0\textbf{(+2.8)}                                              \\
\textbf{InfoCSE}  &   75.8                                  &  77.4(+1.6)                                              \\
\textbf{RankCSE}  &    77.2                                 & 77.5(+0.3)                                              \\
\textbf{TNCSE}    &   \textbf{79.3}                                  &    \textbf{79.6}(+0.3)                                           \\ \hline

\end{tabular}
\captionsetup{width=0.45\textwidth} 
\caption{The impact of adding unlabelled SICKR datasets on model training.}
\label{addingsickr}
\end{center}
\end{table}

\subsection{Impact of Unlabelled SICKR}

We have found that when reproducing SimCSE, ESimCSE, and DiffCSE, we cannot reproduce the results reported in the paper using the official open-source code and default hyperparameters. For example, our reproduction of SimCSE is only about 74\%, which is far from the reported 76.25\%, so we can only improve the reproduction level by adding an unsupervised dataset, based on which we roughly improve the results of SimCSE to about 76\% in order to be fair enough to conduct the subsequent experiments.
In Table \ref{addingsickr}, we report the results we obtained by training the model with the default hyperparameters and the results by adding the SICKR dataset. In addition, we report the results of training TNCSE using the original Wiki1M to demonstrate that adding the unlabelled dataset does not significantly improve the model performance.

\subsection{Discussion: Why Choose Pooler Output Over Last Hidden State?}

In our framework, pooler output is chosen for norm constraints; in this subsection, we discuss the reason.
BERT and RoBERTa, pre-trained models based on the structure of the Transformer
encoder, include two LayerNorms in each of their encoder layers, which are located after the attention layer and after the FNN, respectively, which makes their last hidden state lose the norm feature. No matter the text input to SimCSE-BERT, their last hidden state's norms are almost always distributed between 14 and 16, which cannot do norm constraints.
We remove the last few LayerNorms in the encoder layer of SimCSE-BERT to make the last hidden state get the norm features and apply the Eq. \ref{allloss} on the last hidden state. However, the training results we get are instead worse, and the more LayerNorms we remove, the worse the result is. We consider that this destroys the pre-training information of SimCSE-BERT, which is unfavorable to the TNCSE training.
We report this finding in Figure \ref{discusspdf}. We first randomly select 100 sentences from Wiki1M. Then we remove the last several layers of LayerNorm of the encoder, and report the average norm of the last hidden state, as shown in the bar graph. We train two encoders with the LayerNorm removed using the TNCSE method and report the performance on the STS-B test set. As can be seen from the line graphs, the model almost loses its semantic discriminative ability when all the last six layers of LayerNorm are removed. Overall, the more LayerNorms are preserved, the better the model works, but both are not as good as our proposed scheme employing Pooler output for tensor constraints. The above experiment demonstrates the ingenuity of our design.


\section{Conclusion}

In this work, we propose a new training objective of unsupervised sentence embedding, which can constrain the module length of sentence embedding representations between positive samples; we model it as a tensor's norm and jointly train with ensemble learning. Based on these methods, we propose a new unsupervised sentence embedding representation framework, TNCSE, which becomes a new SOTA method on seven STS tasks.
Moreover, nearly one hundred zero-shot tasks are used to evaluate, and the results show that TNCSE and its derived models outperform other baselines. In addition, we conduct a series of ablation experiments to explore the effects of individual components. 

\section{Acknowledgments}

We thank the anonymous reviewers and area chair for their contributions to this work.

\bibliography{aaai25.bib}

\begin{thebibliography}{29}
\providecommand{\natexlab}[1]{#1}

\bibitem[{Agirre et~al.(2015)Agirre, Banea, Cardie, Cer, Diab, Gonzalez-Agirre, Guo, Lopez-Gazpio, Maritxalar, Mihalcea, Rigau, Uria, and Wiebe}]{15}
Agirre, E.; Banea, C.; Cardie, C.; Cer, D.; Diab, M.; Gonzalez-Agirre, A.; Guo, W.; Lopez-Gazpio, I.; Maritxalar, M.; Mihalcea, R.; Rigau, G.; Uria, L.; and Wiebe, J. 2015.
\newblock {S}em{E}val-2015 Task 2: Semantic Textual Similarity, {E}nglish, {S}panish and Pilot on Interpretability.
\newblock In \emph{Proceedings of the 9th International Workshop on Semantic Evaluation ({S}em{E}val 2015)}, 252--263. Denver, Colorado: Association for Computational Linguistics.

\bibitem[{Agirre et~al.(2014)Agirre, Banea, Cardie, Cer, Diab, Gonzalez-Agirre, Guo, Mihalcea, Rigau, and Wiebe}]{14}
Agirre, E.; Banea, C.; Cardie, C.; Cer, D.; Diab, M.; Gonzalez-Agirre, A.; Guo, W.; Mihalcea, R.; Rigau, G.; and Wiebe, J. 2014.
\newblock {S}em{E}val-2014 Task 10: Multilingual Semantic Textual Similarity.
\newblock In \emph{Proceedings of the 8th International Workshop on Semantic Evaluation ({S}em{E}val 2014)}, 81--91. Dublin, Ireland: Association for Computational Linguistics.

\bibitem[{Agirre et~al.(2016)Agirre, Banea, Cer, Diab, Gonzalez-Agirre, Mihalcea, Rigau, and Wiebe}]{16}
Agirre, E.; Banea, C.; Cer, D.; Diab, M.; Gonzalez-Agirre, A.; Mihalcea, R.; Rigau, G.; and Wiebe, J. 2016.
\newblock {S}em{E}val-2016 Task 1: Semantic Textual Similarity, Monolingual and Cross-Lingual Evaluation.
\newblock In \emph{Proceedings of the 10th International Workshop on Semantic Evaluation ({S}em{E}val-2016)}, 497--511. San Diego, California: Association for Computational Linguistics.

\bibitem[{Agirre et~al.(2012)Agirre, Cer, Diab, and Gonzalez-Agirre}]{sts12}
Agirre, E.; Cer, D.; Diab, M.; and Gonzalez-Agirre, A. 2012.
\newblock {S}em{E}val-2012 Task 6: A Pilot on Semantic Textual Similarity.
\newblock In \emph{*{SEM} 2012: The First Joint Conference on Lexical and Computational Semantics {--} Volume 1: Proceedings of the main conference and the shared task, and Volume 2: Proceedings of the Sixth International Workshop on Semantic Evaluation ({S}em{E}val 2012)}, 385--393. Montr{\'e}al, Canada: Association for Computational Linguistics.

\bibitem[{Agirre et~al.(2013)Agirre, Cer, Diab, Gonzalez-Agirre, and Guo}]{13}
Agirre, E.; Cer, D.; Diab, M.; Gonzalez-Agirre, A.; and Guo, W. 2013.
\newblock *{SEM} 2013 shared task: Semantic Textual Similarity.
\newblock In \emph{Second Joint Conference on Lexical and Computational Semantics (*{SEM}), Volume 1: Proceedings of the Main Conference and the Shared Task: Semantic Textual Similarity}, 32--43. Atlanta, Georgia, USA: Association for Computational Linguistics.

\bibitem[{Cer et~al.(2017{\natexlab{a}})Cer, Diab, Agirre, Lopez-Gazpio, and Specia}]{stsb}
Cer, D.; Diab, M.; Agirre, E.; Lopez-Gazpio, I.; and Specia, L. 2017{\natexlab{a}}.
\newblock {S}em{E}val-2017 Task 1: Semantic Textual Similarity Multilingual and Crosslingual Focused Evaluation.
\newblock In \emph{Proceedings of the 11th International Workshop on Semantic Evaluation ({S}em{E}val-2017)}, 1--14. Vancouver, Canada: Association for Computational Linguistics.

\bibitem[{Cer et~al.(2017{\natexlab{b}})Cer, Diab, Agirre, Lopez{-}Gazpio, and Specia}]{sts17}
Cer, D.~M.; Diab, M.~T.; Agirre, E.; Lopez{-}Gazpio, I.; and Specia, L. 2017{\natexlab{b}}.
\newblock SemEval-2017 Task 1: Semantic Textual Similarity Multilingual and Crosslingual Focused Evaluation.
\newblock In Bethard, S.; Carpuat, M.; Apidianaki, M.; Mohammad, S.~M.; Cer, D.~M.; and Jurgens, D., eds., \emph{Proceedings of the 11th International Workshop on Semantic Evaluation, SemEval@ACL 2017, Vancouver, Canada, August 3-4, 2017}, 1--14. Association for Computational Linguistics.

\bibitem[{Chen et~al.(2022)Chen, Zeynali, Camargo, Fl{\"o}ck, Gaffney, Grabowicz, Hale, Jurgens, and Samory}]{sts22}
Chen, X.; Zeynali, A.; Camargo, C.; Fl{\"o}ck, F.; Gaffney, D.; Grabowicz, P.; Hale, S.; Jurgens, D.; and Samory, M. 2022.
\newblock {S}em{E}val-2022 Task 8: Multilingual news article similarity.
\newblock In Emerson, G.; Schluter, N.; Stanovsky, G.; Kumar, R.; Palmer, A.; Schneider, N.; Singh, S.; and Ratan, S., eds., \emph{Proceedings of the 16th International Workshop on Semantic Evaluation (SemEval-2022)}, 1094--1106. Seattle, United States: Association for Computational Linguistics.

\bibitem[{Chuang et~al.(2022)Chuang, Dangovski, Luo, Zhang, Chang, Soljacic, Li, Yih, Kim, and Glass}]{diffcse}
Chuang, Y.; Dangovski, R.; Luo, H.; Zhang, Y.; Chang, S.; Soljacic, M.; Li, S.; Yih, S.; Kim, Y.; and Glass, J.~R. 2022.
\newblock DiffCSE: Difference-based Contrastive Learning for Sentence Embeddings.
\newblock In Carpuat, M.; de~Marneffe, M.; and Ru{\'{\i}}z, I. V.~M., eds., \emph{Proceedings of the 2022 Conference of the North American Chapter of the Association for Computational Linguistics: Human Language Technologies, {NAACL} 2022, Seattle, WA, United States, July 10-15, 2022}, 4207--4218. Association for Computational Linguistics.

\bibitem[{Conneau and Kiela(2018)}]{senteval}
Conneau, A.; and Kiela, D. 2018.
\newblock SentEval: An Evaluation Toolkit for Universal Sentence Representations.
\newblock In Calzolari, N.; Choukri, K.; Cieri, C.; Declerck, T.; Goggi, S.; Hasida, K.; Isahara, H.; Maegaard, B.; Mariani, J.; Mazo, H.; Moreno, A.; Odijk, J.; Piperidis, S.; and Tokunaga, T., eds., \emph{Proceedings of the Eleventh International Conference on Language Resources and Evaluation, {LREC} 2018, Miyazaki, Japan, May 7-12, 2018}. European Language Resources Association {(ELRA)}.

\bibitem[{Devlin et~al.(2019)Devlin, Chang, Lee, and Toutanova}]{BERT}
Devlin, J.; Chang, M.; Lee, K.; and Toutanova, K. 2019.
\newblock {BERT:} Pre-training of Deep Bidirectional Transformers for Language Understanding.
\newblock In Burstein, J.; Doran, C.; and Solorio, T., eds., \emph{Proceedings of the 2019 Conference of the North American Chapter of the Association for Computational Linguistics: Human Language Technologies, {NAACL-HLT} 2019, Minneapolis, MN, USA, June 2-7, 2019, Volume 1 (Long and Short Papers)}, 4171--4186. Association for Computational Linguistics.

\bibitem[{Gao, Yao, and Chen(2021)}]{gao2021simcse}
Gao, T.; Yao, X.; and Chen, D. 2021.
\newblock SimCSE: Simple Contrastive Learning of Sentence Embeddings.
\newblock In Moens, M.; Huang, X.; Specia, L.; and Yih, S.~W., eds., \emph{Proceedings of the 2021 Conference on Empirical Methods in Natural Language Processing, {EMNLP} 2021, Virtual Event / Punta Cana, Dominican Republic, 7-11 November, 2021}, 6894--6910. Association for Computational Linguistics.

\bibitem[{He et~al.(2023)He, Zhang, Lan, and Zhang}]{iscse}
He, H.; Zhang, J.; Lan, Z.; and Zhang, Y. 2023.
\newblock Instance Smoothed Contrastive Learning for Unsupervised Sentence Embedding.
\newblock In Williams, B.; Chen, Y.; and Neville, J., eds., \emph{Thirty-Seventh {AAAI} Conference on Artificial Intelligence, {AAAI} 2023, Thirty-Fifth Conference on Innovative Applications of Artificial Intelligence, {IAAI} 2023, Thirteenth Symposium on Educational Advances in Artificial Intelligence, {EAAI} 2023, Washington, DC, USA, February 7-14, 2023}, 12863--12871. {AAAI} Press.

\bibitem[{He et~al.(2020)He, Fan, Wu, Xie, and Girshick}]{moco}
He, K.; Fan, H.; Wu, Y.; Xie, S.; and Girshick, R.~B. 2020.
\newblock Momentum Contrast for Unsupervised Visual Representation Learning.
\newblock In \emph{2020 {IEEE/CVF} Conference on Computer Vision and Pattern Recognition, {CVPR} 2020, Seattle, WA, USA, June 13-19, 2020}, 9726--9735. Computer Vision Foundation / {IEEE}.

\bibitem[{Jiang et~al.(2022)Jiang, Jiao, Huang, Zhang, Wang, Zhuang, Wei, Huang, Deng, and Zhang}]{PromptBERT}
Jiang, T.; Jiao, J.; Huang, S.; Zhang, Z.; Wang, D.; Zhuang, F.; Wei, F.; Huang, H.; Deng, D.; and Zhang, Q. 2022.
\newblock PromptBERT: Improving {BERT} Sentence Embeddings with Prompts.
\newblock In Goldberg, Y.; Kozareva, Z.; and Zhang, Y., eds., \emph{Proceedings of the 2022 Conference on Empirical Methods in Natural Language Processing, {EMNLP} 2022, Abu Dhabi, United Arab Emirates, December 7-11, 2022}, 8826--8837. Association for Computational Linguistics.

\bibitem[{Liu et~al.(2023)Liu, Liu, Wang, Wang, Wu, Xian, Zhao, Chen, and Yan}]{rankcse}
Liu, J.; Liu, J.; Wang, Q.; Wang, J.; Wu, W.; Xian, Y.; Zhao, D.; Chen, K.; and Yan, R. 2023.
\newblock RankCSE: Unsupervised Sentence Representations Learning via Learning to Rank.
\newblock In Rogers, A.; Boyd{-}Graber, J.~L.; and Okazaki, N., eds., \emph{Proceedings of the 61st Annual Meeting of the Association for Computational Linguistics (Volume 1: Long Papers), {ACL} 2023, Toronto, Canada, July 9-14, 2023}, 13785--13802. Association for Computational Linguistics.

\bibitem[{Liu et~al.(2019)Liu, Ott, Goyal, Du, Joshi, Chen, Levy, Lewis, Zettlemoyer, and Stoyanov}]{roberta}
Liu, Y.; Ott, M.; Goyal, N.; Du, J.; Joshi, M.; Chen, D.; Levy, O.; Lewis, M.; Zettlemoyer, L.; and Stoyanov, V. 2019.
\newblock RoBERTa: {A} Robustly Optimized {BERT} Pretraining Approach.
\newblock \emph{CoRR}, abs/1907.11692.

\bibitem[{Marelli et~al.(2014)Marelli, Menini, Baroni, Bentivogli, Bernardi, and Zamparelli}]{sickr}
Marelli, M.; Menini, S.; Baroni, M.; Bentivogli, L.; Bernardi, R.; and Zamparelli, R. 2014.
\newblock A {SICK} cure for the evaluation of compositional distributional semantic models.
\newblock In Calzolari, N.; Choukri, K.; Declerck, T.; Loftsson, H.; Maegaard, B.; Mariani, J.; Moreno, A.; Odijk, J.; and Piperidis, S., eds., \emph{Proceedings of the Ninth International Conference on Language Resources and Evaluation, {LREC} 2014, Reykjavik, Iceland, May 26-31, 2014}, 216--223. European Language Resources Association {(ELRA)}.

\bibitem[{Muennighoff et~al.(2023)Muennighoff, Tazi, Magne, and Reimers}]{mteb}
Muennighoff, N.; Tazi, N.; Magne, L.; and Reimers, N. 2023.
\newblock {MTEB:} Massive Text Embedding Benchmark.
\newblock In Vlachos, A.; and Augenstein, I., eds., \emph{Proceedings of the 17th Conference of the European Chapter of the Association for Computational Linguistics, {EACL} 2023, Dubrovnik, Croatia, May 2-6, 2023}, 2006--2029. Association for Computational Linguistics.

\bibitem[{Reimers and Gurevych(2019)}]{sbert}
Reimers, N.; and Gurevych, I. 2019.
\newblock Sentence-{BERT}: Sentence Embeddings using {S}iamese {BERT}-Networks.
\newblock In \emph{Proceedings of the 2019 Conference on Empirical Methods in Natural Language Processing and the 9th International Joint Conference on Natural Language Processing (EMNLP-IJCNLP)}, 3982--3992. Hong Kong, China: Association for Computational Linguistics.

\bibitem[{van~den Oord, Li, and Vinyals(2018)}]{infonce}
van~den Oord, A.; Li, Y.; and Vinyals, O. 2018.
\newblock Representation Learning with Contrastive Predictive Coding.
\newblock \emph{CoRR}, abs/1807.03748.

\bibitem[{Wang and Dou(2023)}]{SNCSE}
Wang, H.; and Dou, Y. 2023.
\newblock {SNCSE:} Contrastive Learning for Unsupervised Sentence Embedding with Soft Negative Samples.
\newblock In Huang, D.; Premaratne, P.; Jin, B.; Qu, B.; Jo, K.; and Hussain, A., eds., \emph{Advanced Intelligent Computing Technology and Applications - 19th International Conference, {ICIC} 2023, Zhengzhou, China, August 10-13, 2023, Proceedings, Part {IV}}, volume 14089 of \emph{Lecture Notes in Computer Science}, 419--431. Springer.

\bibitem[{Wang and Isola(2020)}]{alignment-uniformity}
Wang, T.; and Isola, P. 2020.
\newblock Understanding Contrastive Representation Learning through Alignment and Uniformity on the Hypersphere.
\newblock In \emph{Proceedings of the 37th International Conference on Machine Learning, {ICML} 2020, 13-18 July 2020, Virtual Event}, volume 119 of \emph{Proceedings of Machine Learning Research}, 9929--9939. {PMLR}.

\bibitem[{Wu et~al.(2022{\natexlab{a}})Wu, Gao, Lin, Han, Wang, and Hu}]{infocse}
Wu, X.; Gao, C.; Lin, Z.; Han, J.; Wang, Z.; and Hu, S. 2022{\natexlab{a}}.
\newblock InfoCSE: Information-aggregated Contrastive Learning of Sentence Embeddings.
\newblock In Goldberg, Y.; Kozareva, Z.; and Zhang, Y., eds., \emph{Findings of the Association for Computational Linguistics: {EMNLP} 2022, Abu Dhabi, United Arab Emirates, December 7-11, 2022}, 3060--3070. Association for Computational Linguistics.

\bibitem[{Wu et~al.(2022{\natexlab{b}})Wu, Gao, Zang, Han, Wang, and Hu}]{esimcse}
Wu, X.; Gao, C.; Zang, L.; Han, J.; Wang, Z.; and Hu, S. 2022{\natexlab{b}}.
\newblock ESimCSE: Enhanced Sample Building Method for Contrastive Learning of Unsupervised Sentence Embedding.
\newblock In Calzolari, N.; Huang, C.; Kim, H.; Pustejovsky, J.; Wanner, L.; Choi, K.; Ryu, P.; Chen, H.; Donatelli, L.; Ji, H.; Kurohashi, S.; Paggio, P.; Xue, N.; Kim, S.; Hahm, Y.; He, Z.; Lee, T.~K.; Santus, E.; Bond, F.; and Na, S., eds., \emph{Proceedings of the 29th International Conference on Computational Linguistics, {COLING} 2022, Gyeongju, Republic of Korea, October 12-17, 2022}, 3898--3907. International Committee on Computational Linguistics.

\bibitem[{Yan et~al.(2021)Yan, Li, Wang, Zhang, Wu, and Xu}]{consert}
Yan, Y.; Li, R.; Wang, S.; Zhang, F.; Wu, W.; and Xu, W. 2021.
\newblock ConSERT: {A} Contrastive Framework for Self-Supervised Sentence Representation Transfer.
\newblock In Zong, C.; Xia, F.; Li, W.; and Navigli, R., eds., \emph{Proceedings of the 59th Annual Meeting of the Association for Computational Linguistics and the 11th International Joint Conference on Natural Language Processing, {ACL/IJCNLP} 2021, (Volume 1: Long Papers), Virtual Event, August 1-6, 2021}, 5065--5075. Association for Computational Linguistics.

\bibitem[{Zhang et~al.(2022)Zhang, Zhu, Wang, Xu, Li, and Zhao}]{arccse}
Zhang, Y.; Zhu, H.; Wang, Y.; Xu, N.; Li, X.; and Zhao, B. 2022.
\newblock A Contrastive Framework for Learning Sentence Representations from Pairwise and Triple-wise Perspective in Angular Space.
\newblock In \emph{Proceedings of the 60th Annual Meeting of the Association for Computational Linguistics (Volume 1: Long Papers)}, 4892--4903. Dublin, Ireland: Association for Computational Linguistics.

\bibitem[{Zhuo et~al.(2023)Zhuo, Sun, Wang, Zhu, and Yang}]{whencse}
Zhuo, W.; Sun, Y.; Wang, X.; Zhu, L.; and Yang, Y. 2023.
\newblock WhitenedCSE: Whitening-based Contrastive Learning of Sentence Embeddings.
\newblock In Rogers, A.; Boyd{-}Graber, J.~L.; and Okazaki, N., eds., \emph{Proceedings of the 61st Annual Meeting of the Association for Computational Linguistics (Volume 1: Long Papers), {ACL} 2023, Toronto, Canada, July 9-14, 2023}, 12135--12148. Association for Computational Linguistics.

\bibitem[{Zong and Zhang(2023)}]{zong}
Zong, T.; and Zhang, L. 2023.
\newblock An Ensemble Distillation Framework for Sentence Embeddings with Multilingual Round-Trip Translation.
\newblock In \emph{Proceedings of the AAAI Conference on Artificial Intelligence}, volume~37, 14074--14082.

\end{thebibliography}

\end{document}